\documentclass[a4paper,fleqn]{cas-dc}

\usepackage[authoryear,longnamesfirst]{natbib}
\usepackage{amsmath}
\usepackage{amssymb}
\usepackage{CJKutf8}
\usepackage{makecell}
\usepackage{booktabs}
\usepackage{tabularx}

\usepackage{algorithm}
\usepackage{algorithmic}

\def\tsc#1{\csdef{#1}{\textsc{\lowercase{#1}}\xspace}}
\tsc{WGM}
\tsc{QE}

\begin{document}
\let\WriteBookmarks\relax
\def\floatpagepagefraction{1}
\def\textpagefraction{.001}

\shorttitle{Pre-training Language Model Incorporating Domain-specific Heterogeneous Knowledge into A Unified Representation}    

\shortauthors{Zhu, Peng and Lyu et al.}  

\title [mode = title]{Pre-training Language Model Incorporating Domain-specific Heterogeneous Knowledge into A Unified Representation}  



%






\credit{<Credit authorship details>}

\affiliation[1]{organization={Department of Computer Science and Technology, Tsinghua University},
city={Beijing},
country={China}}
\affiliation[2]{organization={Huawei Noah’s Ark Lab},
city={Beijing},
country={China}}
\author[1]{Hongyin Zhu}
\ead{zhuhongyin2020@mail.tsinghua.edu.cn}
\author[1]{Hao Peng}
\ead{h-peng17@mails.tsinghua.edu.cn}
\author[1]{Zhiheng Lyu}
\ead{lvzh18@mails.tsinghua.edu.cn}
\author[1]{Lei Hou}
\ead{houlei@tsinghua.edu.cn}
\author[1]{Juanzi Li}\cormark[1]
\ead{lijuanzi@tsinghua.edu.cn; Phone: +86 010-62781461; Postal address: Room 10-205, East Main Building, Tsinghua University, No. 30 Shuangqing Road, Haidian District, Beijing, 100084}
\author[2]{Jinghui Xiao}
\ead{xiaojinghui4@huawei.com}






\cortext[1]{Corresponding author}



\begin{abstract}
Existing technologies expand BERT from different perspectives, e.g. designing different pre-training tasks, different semantic granularities, and different model architectures. Few models consider expanding BERT from different text formats. In this paper, we propose a heterogeneous knowledge language model (\textbf{HKLM}), a unified pre-trained language model (PLM) for all forms of text, including unstructured text, semi-structured text, and well-structured text. To capture the corresponding relations among these multi-format knowledge, our approach uses masked language model objective to learn word knowledge, uses triple classification objective and title matching objective to learn entity knowledge and topic knowledge respectively. To obtain the aforementioned multi-format text, we construct a corpus in the tourism domain and conduct experiments on 5 tourism NLP datasets. The results show that our approach outperforms the pre-training of plain text using only 1/4 of the data. We further pre-train the domain-agnostic HKLM and achieve performance gains on the XNLI dataset.
\end{abstract}




\maketitle

\section{Introduction}
The expansion of PLMs aims to augment the existing PLMs to better understand the downstream text. These methods design different learning tasks (\cite{DBLP:journals/corr/abs-1904-09223}), different semantic granularities (\cite{DBLP:conf/acl/ZhangHLJSL19}), different model architectures (\cite{DBLP:conf/acl/DaiYYCLS19}), and different learning algorithms (\cite{DBLP:conf/iclr/ClarkLLM20}) but they usually use single-format text to pre-train the model and lack the learning of document structure and relevant knowledge. 
There is a large amount of unused semi-structured and well-structured text. Together with unstructured free text, we refer to them as multi-format text. These data are essential for the hierarchical understanding of words, entities, and paragraphs. 
This leads us to ask whether we can utilize multi-format text in pre-training. 

\begin{figure}[!h]
\centering
\includegraphics[width=3in]{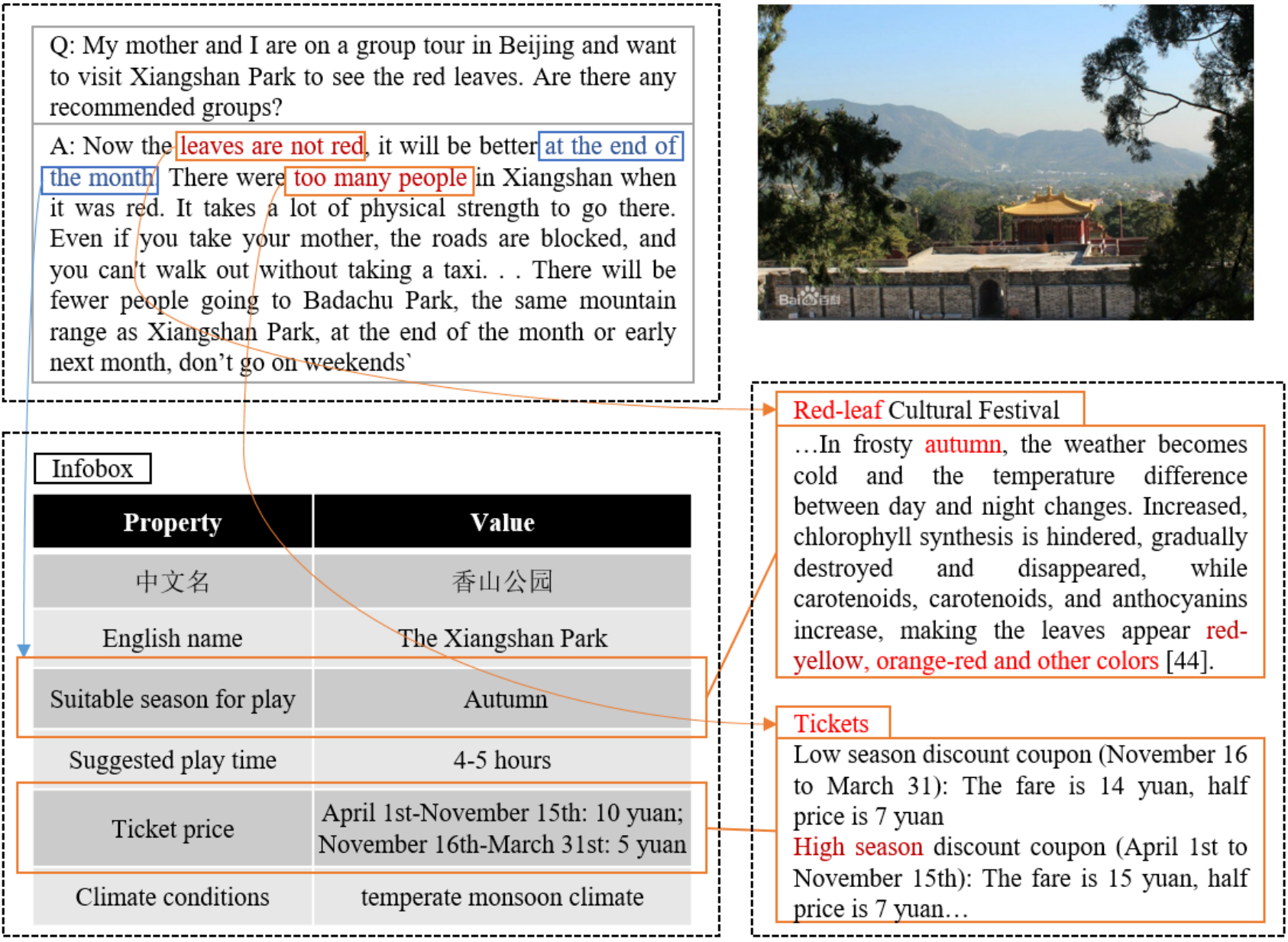}
\caption{An example to demonstrate the usage of unstructured, semi-structured, and well-structured knowledge}
\label{example.fig}
\end{figure}

First, let us take Figure \ref{example.fig} as an example to illustrate the multi-format text resources. For the question, ``\textit{My mother and I are on a group tour ...}'', although the scenery described in the candidate answer does not mention the name of the tourist attraction, with the help of the Baidu encyclopedia, we can still understand that the answer is describing the scenery of Xiangshan Park and claiming that it is not the best time to travel. The key words ``\textit{leaves are not red}'', ``\textit{at the end of the month}'', ``\textit{too many people}'' have corresponding semi-structured subsection titles ((sub)headings), paragraphs and well-structured knowledge triples (\textit{Xiangshan Park, suitable season for play, autumn}) in the encyclopedia webpage. The paragraphs also correspond to relevant knowledge triples.

Most prior BERT (\cite{DBLP:conf/naacl/DevlinCLT19}) expansion works are designed to use plain text and entity type or change the model architecture, rarely considering semi-structured text and well-structured text in Figure \ref{example.fig}. Beltagy et al. (\cite{DBLP:conf/emnlp/BeltagyLC19}) use scientific publications to further pre-train the SCIBERT to improve performance on downstream scientific NLP tasks. Xiong et al. (\cite{DBLP:conf/iclr/XiongDWS20}) propose type-constrained entity replacement pre-training task for knowledge learning. THU-ERNIE (\cite{DBLP:conf/acl/ZhangHLJSL19}) proposes to incorporate the contextual representations with separate knowledge graph (KG) embeddings, but they do not consider the correlation between knowledge triples and the text. Semi-structured text is important for language understanding, which has been proven effective in question answering (QA) systems (\cite{min2020neurips,kwiatkowski2019natural}), but few works explicitly model semi-structured text into the pre-training process. This highlights the need to model the document structure, relevant knowledge triples, and plain text in the same representation space. 

Modeling multi-format text is a nontrivial task. The main difficulty lies in finding the correspondence between heterogeneous knowledge resources. Our goal is to find an effective way to model unstructured paragraphs, semi-structured titles, and well-structured knowledge triples and let them interact with each other. Using the relationship between the titles and paragraphs, the model can understand the topic of the paragraph. Inspired by the use of knowledge triples (\cite{DBLP:journals/ibmrd/Chu-CarrollFBCSW12}), these triples help add explicit restrictions or complement information to the text, evaluate the information expressed and improve the interpretability.

We propose a heterogeneous knowledge language model (HKLM) that simultaneously models unstructured, semi-structured, and well-structured text into the same contextual representation space. To obtain the aforementioned multi-format text, we construct a corpus in the tourism domain and pre-train our \textbf{TravelBERT}. Specifically, our multi-format text comes from the Baidu encyclopedia webpages of Chinese Tourist Attractions, since entity-oriented resources have rich aligned heterogeneous knowledge. The HKLM is suitable for training using encyclopedia articles. At the same time, we also use a plain text corpus of travel guides to pre-train another version, like SCIBERT. The unstructured text has a larger amount of data (4 times in the experiment) than encyclopedia articles. We combine three objective functions to jointly pre-train the multi-format text. For unstructured text, we adopt the masked language model (MLM) objective to train the domain adaption model. For semi-structured text, we propose title matching training (TMT) to classify whether the title matches the paragraph. For well-structured text, we propose a triple classification (TC) task to classify whether the knowledge triple is modified. To align the knowledge triples with the plain text, we use a heuristic search method to calculate the similarity between the text and the triples.

We evaluate the model using 5 downstream tourism NLP tasks, including named entity recognition (NER), open information extraction (IE), question answering (QA), fine-grained entity typing (ET), and dialogue. Our method achieves significant improvements on multiple datasets. We further pre-train the domain-agnostic HKLM ({\bf HKBERT}) with 20 million Baidu encyclopedia webpages and achieve performance improvements on the XNLI dataset. Source code for pre-training and downstream tasks as well as datasets and knowledge graphs are available on Github\footnote{https://github.com/liftkkkk/travelbert}. The main contributions of this paper are as follows:

(1) We propose to model heterogeneous knowledge into a unified representation space with different objectives, thereby allowing them to interact with each other.

(2) We pre-train the TravelBERT with two schemes, using plain text and the proposed HKLM. Experiments show that using 1/4 of the plain text, the HKLM outperforms the pre-training of plain text.

(3) We further pre-train HKBERT on the general domain and demonstrate performance improvements on the XNLI dataset. 

(4) We construct 4 datasets for evaluating downstream tourism NLP tasks.

The paper is organized as follows: Section \ref{section2} presents a literature review on domain-specific pre-training, knowledge-aware pre-training, and NLP tasks utilizing document structure. Section \ref{section3} details the proposed framework HKBERT for modeling multi-format text into a unified representation, as well as methods for handling downstream tasks. Section \ref{section4} presents the experimental results and analysis in the tourism domain and general domain. Finally, Section \ref{section5} presents the conclusions of this study.
\section{Related Work}
\label{section2}
\subsection{Domain-specific Pre-training}
Fine-tuning large PLMs (\cite{DBLP:conf/naacl/DevlinCLT19,DBLP:conf/nips/BrownMRSKDNSSAA20,radford2019language,radford2018improving}) have achieved state-of-the-art results in downstream NLP tasks. Continued pre-training PLMs on a large corpus of unlabeled domain-specific text is helpful to the domain of a target task.  Gururangan et al. (\cite{DBLP:conf/acl/GururanganMSLBD20}) investigate the impact of domain-adaptive pre-training and tasks-adaptive pre-training on language models. Further research on the domain adaptability of 
PLMs has become a promising topic. For the scientific domain, there are SCIBERT (\cite{DBLP:conf/emnlp/BeltagyLC19}) and PatentBERT (\cite{lee2019patentbert}). For the biomedical domain, there are BioBERT (\cite{DBLP:journals/bioinformatics/LeeYKKKSK20}), extBERT (\cite{tai2020exbert}), PubMedBERT (\cite{DBLP:journals/corr/abs-2007-15779}), ClinicalBERT (\cite{DBLP:journals/corr/abs-1904-05342}), MT-ClinicalBERT (\cite{DBLP:journals/corr/abs-2004-10220}). For the financial domain, there are FinBERT (\cite{DBLP:journals/corr/abs-1908-10063,DBLP:conf/ijcai/0001HH0Z20}). Rongali et al. (\cite{DBLP:journals/corr/abs-2004-02288}) propose to mitigate catastrophic forgetting during domain-specific pre-training. However, most studies only use the plain text corpus without considering the document structure and structured knowledge, which leads to the loss of important information in learning. Our proposed method further incorporates semi-structured and well-structured text.

\subsection{Knowledge-aware Pre-training}
Knowledge-aware pre-training is designed to expand text input and introduce external knowledge resources into PLMs, rather than only considering the input text. Xiong et al. (\cite{DBLP:conf/iclr/XiongDWS20}) propose WKLM to use the type-constrained entity replacement for knowledge learning, but they do not consider other attributes of entities. 
Liu et al. (\cite{DBLP:conf/aaai/LiuZ0WJD020}) propose K-BERT which computes the attention score between tokens and KG triples. This method changes the sentence structure by inserting triples in the fine-tuning stage. Yao et al. (\cite{DBLP:journals/corr/abs-1909-03193}) propose KG-BERT, and they let BERT classify whether the triple is valid during the pre-training process, and then apply it to the knowledge graph completion task. Wang et al. (\cite{DBLP:journals/corr/abs-1911-06136}) propose KEPLER that encodes entity description as entity embedding to jointly train the knowledge embeddings and masked language model. ERNIE 1.0 (\cite{DBLP:journals/corr/abs-1904-09223}) propose the whole word masking to mask whole word, entity, and phrase. This method only considers the entity mention and sentence span. Cui et al. (\cite{chinese-bert-wwm}) propose 3 different masking strategies, including whole word masking, N-gram masking, MLM as correction masking, which they claim can force masked language models to recover whole words and mitigate the differences between the pre-training and fine-tuning stages. 

THU-ERNIE (\cite{DBLP:conf/acl/ZhangHLJSL19}) proposes the information fusion layer for the mutual integration of words and entities.  LUKE (\cite{DBLP:conf/emnlp/YamadaASTM20}) proposes entity-aware self-attention and masks tokens and entities in pre-training. TaBERT (\cite{DBLP:conf/acl/YinNYR20}) proposes to learn the joint representations of textual and tabular data. We consider the alignment of multi-format text and simultaneously model them in the same representation space. 
Wang et al. (\cite{DBLP:conf/acl/WangTDWHJCJZ21}) claim that many knowledge injection methods lead to the dilution of past knowledge. They propose the K-ADAPTER framework to inject factual and linguistic knowledge into RoBERTa (\cite{DBLP:journals/corr/abs-1907-11692}) through two adapters. An adapter is a specific knowledge model inserted into a pre-trained model whose input is the hidden state of the middle layers. Petroni et al. (\cite{DBLP:conf/emnlp/PetroniRRLBWM19}) claim that the pre-trained language model can store the relational knowledge existing in the training data while learning language knowledge, and may be able to answer questions structured as a cloze test, they propose language model analysis probe LAMA. Pre-trained language models can recall factual knowledge without fine-tuning. Sun et al. (\cite{DBLP:journals/corr/abs-2107-02137}) propose a pre-training framework ERNIE 3.0, which fuses auto-regressive networks and auto-encoder networks, and the trained model can handle natural language understanding and generation tasks through zero-shot learning, few-shot learning, or fine-tuning. A 4TB corpus of plain text and knowledge graphs is used to pre-train a large-scale knowledge augmentation model with 10B parameters.

\subsection{NLP Tasks Utilizing Document Structure}
Document structure (\cite{power2003document}) describes the organization of a document into graphical constituents like sections, paragraphs, sentences, bulleted lists, and figures. Document structure mainly represents two levels: logical structure (such as outline, sections, etc.) and visual structure (such as element layout, font size, color, etc.). Document structure provides more information than sentences. Currently, it has become a popular trend to move research from sentence level to document level in many fields, such as DocRED (\cite{DBLP:conf/acl/YaoYLHLLLHZS19}), QA (\cite{DBLP:conf/emnlp/WangSGDJ20}), etc. Top-performing systems of EfficientQA (\cite{min2020neurips,kwiatkowski2019natural}) prove that considering infobox, lists, and tables on Wikipedia webpages can lead to performance gains because of the explicit information. However, the structured information is only used as a candidate for reading comprehension. 
Lockard et al. (\cite{DBLP:conf/acl/LockardSDH20}) propose to encode visual elements including layout, font size, and color in the graph attention network to improve the performance of relation extraction on webpages. The document structure is mainly used as an input feature for downstream tasks. Different from their methods, we introduce the document structure in the pre-training process, enabling the model to learn the topic knowledge of paragraphs. 

\section{Methods}
\label{section3}
\begin{figure*}[!h]
\centering
\includegraphics[width=\linewidth]{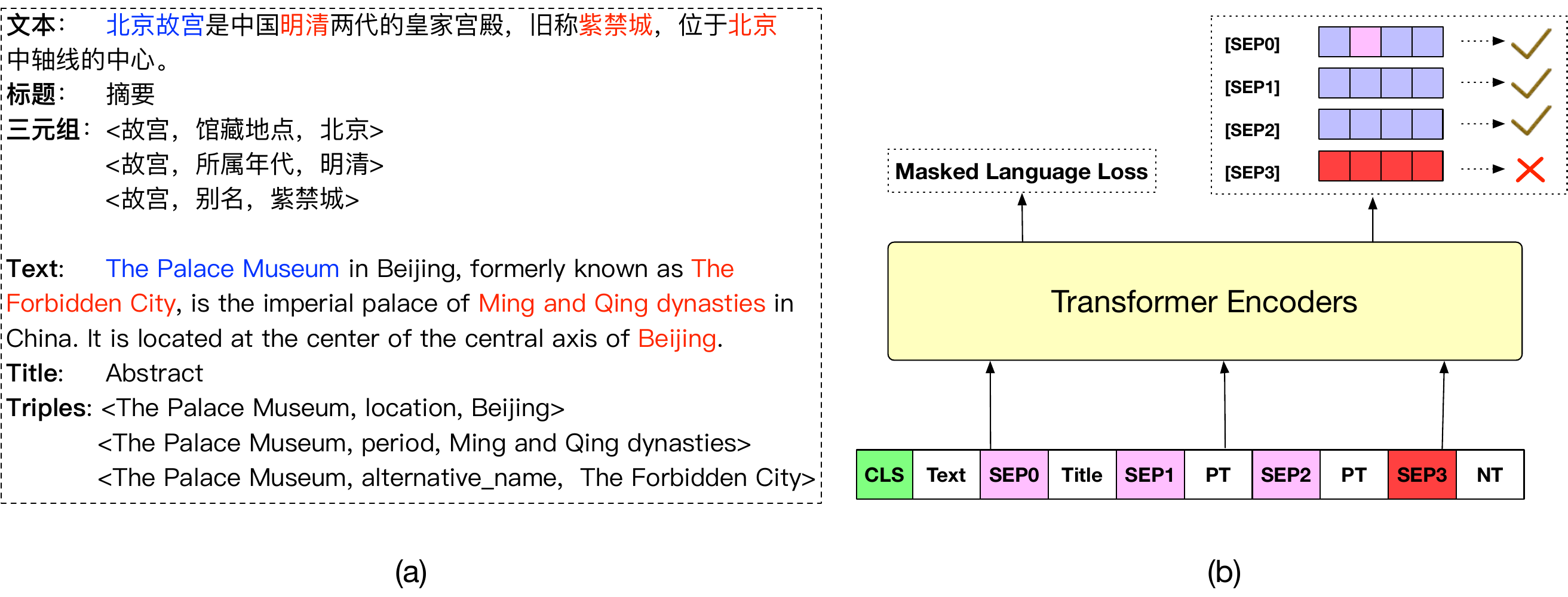}
\caption{Schematic diagram of heterogeneous knowledge language model}
\label{kast.fig}
\end{figure*}
This section explains the mechanism of the proposed HKLM. Suppose we have a document set $D$, a title set $T$, and a knowledge base $KG$. For \{$s^{(e)} \mid s^{(e)} \in D^{(e)}$\}, \{$t^{(e)} \mid t^{(e)} \in T^{(e)}$\} and \{$kg^{(e)} \mid kg^{(e)} \in KG^{(e)}$\}, the superscript $^{(e)}$ denotes the entity-oriented resource. As shown in Figure \ref{kast.fig}, for the entity $e$ ``\textit{The Palace Museum}'', given a piece of text description $s^{(e)}$:$\{w_1,w_2,...,w_l\}$ ``\textit{The Palace Museum in Beijing ....}'', corresponding title $t^{(e)}$ ``\textit{Abstract}'' and relevant knowledge triples $kg^{(e)}$:$\{(e,p_1,o_1),...,(e,p_k,o_k)\}$ ``\textit{(The Palace Museum, location, Beijing),...}'', our goal is to learn a PLM to incorporate the knowledge of unstructured text, semi-structured text and well-structured text.

Our main effort lies in designing an unsupervised pre-training method to augment the contextual representation by leveraging unstructured paragraphs, semi-structured titles, and well-structured knowledge triples. The main improvements consist of two aspects, injecting entity knowledge and topic knowledge. In the following, we first describe the process of pre-training TravelBERT using plain text. Then, we describe the proposed HKLM and its training method. 

\subsection{Pre-training Language Models with Unstructured Text}

Howard et al. (\cite{DBLP:conf/acl/RuderH18}) show that further pre-training a language model on a target domain corpus improves the eventual classification performance. We use the unstructured text in the tourism domain to further pre-train BERT.

Each text sequence is concatenated with special symbols, classification [CLS] and separator [SEP], denoted as $\langle\mbox{[CLS]};s;\mbox{[SEP]}\rangle$.
\begin{align}
{\bf h} = &\mathcal{F}_{bert}(\langle\mbox{[CLS]};s;\mbox{[SEP]}\rangle) 
\end{align}
where ${\bf h} \in \mathbb{R}^{d\times l}$ is the representation of each token. $s$ is the unstructured text. $d$ and $l$ are the dimension and sequence length. $\mathcal{F}_{bert}(\cdot)$ denotes the network defined in (\cite{DBLP:conf/naacl/DevlinCLT19}). Then we use the MLM objective to train the model. Given that BERT is a representative PLM, all studies in this paper use BERT as the backbone. 

\subsection{Heterogeneous Knowledge Language Model}
We care about whether the multi-format text data source can be widely shared by different models. The encyclopedia article generated from collective intelligence is a kind of multi-format text data on hand. Many studies focus on the free text when using encyclopedia articles, ignoring the rich document elements. We aim to use the document structure and infobox triples. Besides, the semi-structured text and internal links in encyclopedia webpages can also be used for data annotation in downstream NLP tasks. Next, we will introduce the way we model multi-format text. The core of pre-training multi-format text is to align them so that different textual modalities can interact. The challenge is to maintain the alignment of multi-format text when the document is divided into many fragments. 

As shown in Figure \ref{kast.fig} (a), the original data consists of free text, corresponding titles, and relevant knowledge triples. We use the Chinese data to train our model. For convenience, we display the data in Chinese and English. As shown in the lower part of Figure \ref{kast.fig} (b), the input is composed of the \textbf{Text} ``\textit{The Palace Museum in Beijing ....}'', \textbf{Title} \textit{Abstract}, positive triple \textbf{PT} and negative triple \textbf{NT} generated by modifying the original triple, denoted as $\langle\mbox{[CLS]};s^{(e)};\mbox{[SEP0]};t^{(e)};\mbox{[SEP1]};(e, p_j,o_i);$ $\mbox{[SEP2]};...\rangle$. We add a new symbol [SEP0] to identify the title. [SEPi] $(i>0)$ is used to represent each knowledge triple. Each element of the triple is treated as text rather than an identifier. The advantage is that different forms of knowledge can be represented in the same contextual representation space. The downside is the lack of linkage and disambiguation between knowledge triples. We use formula \eqref{allrep} to compute the representation of each element. 
\begin{align}
\label{allrep}
&{\bf h}^{(D)},{\bf h}^{(T)},{\bf h}^{(KG)} = \mathcal{F}_{bert}(\langle\mbox{[CLS]};s^{(e)}; \mbox{[SEP0]}; \\ \nonumber & t^{(e)};\mbox{[SEP1]};(e, p_j,o_i);\mbox{[SEP2]}...\rangle) 
\end{align}
This model needs to predict whether the \textbf{Title} matches the \textbf{Text} and whether the predicate of the triple is modified. Meanwhile, we retain the MLM loss to help the model learn from the text in the tourism domain. Next, we will describe the model input and training mechanism in detail.

\subsubsection{Learning Entity Knowledge Through Well-structured Text}
The well-structured text refers to the text organized according to a certain pattern. In this paper, the well-structured text represents the knowledge triples of the infobox. Knowledge triples can directly and centrally describe the attributes of the entity. This allows the model to learn entity information that is unclear or difficult to capture in the context. Such information is important for enhancing entity-oriented downstream tasks. Well-structured text can provide knowledge guidance for understanding the free text. By introducing rich attributes and relations of entities, the representation of entities in the text is naturally enriched.

Baidu-ERNIE (\cite{DBLP:journals/corr/abs-1904-09223}) masks the entities or phrases in the sentence so that the model can better consider the entity's context. Our model randomly masks tokens with a probability of 15\%. To better learn the attributes of entities, in addition to masking the free text, our model also randomly masks the knowledge triples. Usually, an entity may have several to dozens of attributes, such as \textit{address, location, famous scenery, climate type, attraction level, type of attraction, building time, complete time}, etc. To reduce the computational pressure and retain sufficient free text, we adopt a knowledge triple retrieval strategy to use the relevant knowledge triples corresponding to the free text. We use TF-IDF similarity (\cite{tata2007estimating}) for control and only use the knowledge triples contained in text descriptions. Then we use TF-IDF vectors to calculate the similarity of each triple with the text, as shown below. 
\begin{align}
similarity = cos(vec(s^{(e)}), vec((e,p,o)))
\end{align}
where $vec(\cdot)$ is the vectorization (mapping) function to convert text to TF-IDF vector. $s^{(e)}$ and $(e,p,o)$ are seen as different documents. Then we calculate the cosine similarity.

Specifically, we calculate the TF-IDF of the corpus composed of all the Baidu encyclopedia webpages of Chinese tourist attractions and knowledge triples, as shown below. Each triple is also seen as a text sequence.
\begin{align}
\hbox{TF-IDF} = \frac{f_{t,d}}{\sum_{t^{'}\in d}f_{t^{'},d}} \times \log\frac{N}{n_t}
\end{align}
where $d$ is the document that contains term $t$. $f_{t,d}$ is the term frequency of $t$ in document $d$. $N$ is the number of documents in the corpus. $n_t$ is the number of documents that contains $t$. We observe that 73\% of the text samples can be paired with at least one triple.

we adopt a heuristic method in the data processing of triples to only retain some triples that are closely related to unstructured text. The advantage of this method is that it can increase the relatedness of triples and unstructured text, ensure that there is enough space for unstructured text, and reduce computational pressure. Its disadvantage is that the total amount of triples is sometimes not enough. Our motivation for doing this is to focus on the generality of this method. When we are faced with larger-scale triples, such as open relation extraction results. When we face large-scale triples, such as knowledge graphs or knowledge extracted with open relations, our method can extract the most relevant triples without modification. We use this approach to address the limitations of the model.

Usually, training knowledge graph embeddings (such as TransE (\cite{DBLP:conf/nips/BordesUGWY13})) requires optimizing objective function of $h+r \approx t$ to learn the internal relation, where $h$, $r$ and $t$ denote the head entity, relation and tail entity respectively. When we inject knowledge triples as text, we hope to establish a natural connection between free text and well-structured text. The challenge is that the knowledge triples and free text exist separately, and this association has not been annotated before. Ideally, one option is to train the model through the task of generating triples from free text. However, this training task will increase the complexity of the model. Finally, we simplify it to the process of constructing triple-paragraph pairs using TF-IDF. We use a program to add noise to the triples, and the model only needs to predict whether each triple is modified. The model can use unstructured text to validate the knowledge triples.

We call it a triple classification (TC) task which is designed to let the model classify whether each triple is modified. The method of adding noise is the attribute resampling operation, which is designed to randomly replace some triples' attributes $p$ with other triples' attributes $\hat{p}$, as shown in formula \eqref{replace}. Resampling attributes can improve the model's understanding of relational semantics. Conversely, resampling attribute values may cause the model to fail to classify confusing results (e.g. numeric attribute values), thereby increasing the false positive rate of model predictions.

\begin{align}
\label{replace}
\hat{p} \sim \text{Uniform}(P\backslash\{p\})
\end{align}
where $P$ denotes all attributes in the KG. 
We sample predicates from the same Baidu encyclopedia webpage for replacement. Uniform sampling is more common and unrestricted. The disadvantage is the lack of consideration of the semantics of the candidate predicate in the sampling process. In future research, we will improve the sampling quality by designing semantic-based sampling algorithms. 

\subsubsection{Learning Paragraph Knowledge Through Semi-structured Text}
The MLM focuses on word prediction but lacks the concept of paragraphs and subsections. Pre-training for understanding the overall topic of the paragraph is a nontrivial task. In the unsupervised scenario, we use the titles of the document as the natural label of corresponding paragraphs to guide the model to understand the topic of the paragraph. However, the titles of paragraphs cannot be enumerated, so they are not feasible as classes. Adding a generative model will increase the complexity of our TravelBERT. To make this method more versatile, we use a program to automatically construct the title-paragraph pairs, and the model only needs to predict whether the title is modified based on the paragraph. 

Inspired by the use of the relationship between the title and the content of title-oriented documents in IBM Watson DeepQA (\cite{DBLP:journals/ibmrd/Chu-CarrollFBCSW12}), we model the semi-structured text by proposing title matching training (TMT). TMT aims to classify whether the title matches the paragraph. For multi-level titles, we use the title closest to the paragraph to get a more specific topic. When pre-training the language model, we concatenate the text with the corresponding titles, denoted as $\langle\mbox{[CLS]};s^{(e)};\mbox{[SEP0]};t^{(e)}\rangle$. We use the representation of [SEP0] for classification.

To generate negative samples, for the title $t^{(e)}$ ``\textit{Abstract}'', our method will sample another title $\hat{t}^{(e)}$ ``\textit{History}'' in the same article to replace the original title, as shown below.
\begin{align}
\label{replace2}
\hat{t}^{(e)} \sim \text{Uniform}(T^{(e)}\backslash\{t^{(e)}\})
\end{align}

Our motivation for proposing this approach is that we want to train an encoder model, not a generative or encoder-decoder model. On the one hand, generative models or encoder-decoder models are not suitable for many downstream tasks, such as NER, FET, IR-QA, open relation extraction, etc. On the other hand, when we only use the encoder module (BERT) to solve downstream tasks, the knowledge of the decoder module is missing.

\subsection{Model Training}
For the three tasks, we can optimize the combined objective function. 
\begin{align}
\label{objective} 
\min\limits_{\Theta}\mathcal{L}&=\sum_{i=1}^{\mid D\mid}(\mathcal{L}_{i}^{(mlm)}+ \mathcal{L}_{i}^{(tc)} + \mathcal{L}_{i}^{(tmt)})
\end{align}
where $\mathcal{L}_{i}^{(mlm)}$, $\mathcal{L}_{i}^{(tc)}$ and $\mathcal{L}_{i}^{(tmt)}$ are the objectives of three tasks respectively. $\mid D\mid$ is the size of the dataset. $\Theta$ is the model parameter. The training loss is to sum the deviation of the cloze task, the deviation of triple classification, and the deviation of title matching. We adopt the negative log-likelihood as the objective, as shown below. 
\begin{align}
\mathcal{L}_{}^{(mlm)}&=-\sum_{i} \log p^{(mlm)}(y_i^{(mlm)}\mid{\bf h}^{(D)};\Theta) 
\end{align}
where $p^{(mlm)}(\cdot)$ represents the probabilities of the true class of the cloze test. $i$ denotes the indices of the token. 
\begin{align}
\mathcal{L}_{}^{(tc)}&=-\sum_{j} \log p^{(tc)}(y_j^{(tc)}\mid{\bf h}^{(KG)};\Theta)) 
\end{align}
where $p^{(tc)}(\cdot)$ represents the probabilities of the true class of triple classification tasks. $j$ denotes the indices of triple. 
\begin{align}
\mathcal{L}_{}^{(tmt)}&=-\log p^{(tmt)}(y^{(tmt)}\mid{\bf h}^{(T)};\Theta)
\end{align}
where $p^{(tmt)}(\cdot)$ represents the probabilities of the true class of title matching task. 
\begin{algorithm}[h]
\caption{The pre-training algorithm}
\label{bpl.alg}
\hspace*{\algorithmicindent} \textbf{Input}: Aligned multi-format text $samples$, batch size $bs$ \\
\hspace*{\algorithmicindent} \textbf{Output}: Loss of 3 subtasks

\begin{algorithmic}[1]
\STATE Initialize random number generator $p_1$, $p_2$
\STATE $batch$ = []
\FOR{ $ j \leftarrow 0,samples.length-1 $}
  \IF {$p_1$ > 0.5}
    \STATE Replace title (equation \eqref{replace2}) and revise the label
  \ENDIF
  \IF {$p_2$ > 0.5}
    \STATE Revise triples (equation \eqref{replace}) and add labels
  \ENDIF
  \STATE Mask tokens and take missing words as labels
  \STATE $batch$.add([(masked tokens, token labels), (title, title\_label), (triples, triple\_label)])
  \IF {$batch$.length \% $bs$ == 0}
    \STATE Get ${\bf h}^{(D)},{\bf h}^{(T)},{\bf h}^{(KG)} = \mathcal{F}_{bert}(batch)$ 
    \STATE $batch$ = []
    \STATE Calculate the loss function for 3 subtasks
  \ENDIF

\ENDFOR
\end{algorithmic}
\end{algorithm}

As shown in Algorithm 1, this model first initializes the random number generator in line 1. Then, this model iterates over all samples during model training. In lines 4-6, this model randomly modifies the title based on equation \eqref{replace2}. In lines 7-9, this model randomly modifies triples based on equation \eqref{replace}. This model randomly masks tokens in line 10. In line 11, we collect samples to form the batch input. In line 13, this model generates the BERT representation. In line 15, this model calculates the loss function for the 3 subtasks. Then, we can update the model parameters using an optimization algorithm.

We briefly discuss the asymptotic complexity of our approach. For simplicity, we assume all hidden dimension is $\rho$ and that the complexity of matrix($\rho \times \rho$)-vector($\rho \times 1$) multiplication is $O(\rho^2)$. Calculating the BERT representation of takes $O(C_{bert})$. To randomly sample from $n$ triples and $m$ titles requires a random sampling operation, which takes $O(m+n)$. Computing the loss function requires a linear layer to reduce the vector dimension, which requires $O(k\rho^2)$. Thus, the total complexity is $O(C_{bert}+k\rho^2+n)$. BERT mainly consists of matrix-vector multiplication, so it takes $O(ml\rho^2)$ with optimized calculation, where $m$ is the number of matrix-vector multiplication and $l$ is the sequence length. The proposed method is still close to the computational complexity of BERT. Our proposed method does not change the computational cost of BERT, so the complexity of the method is low.

\subsection{Fine-tuning TravelBERT for Tourism NLP Tasks}

\begin{figure*}[!h]
\centering
\includegraphics[width=0.99\linewidth]{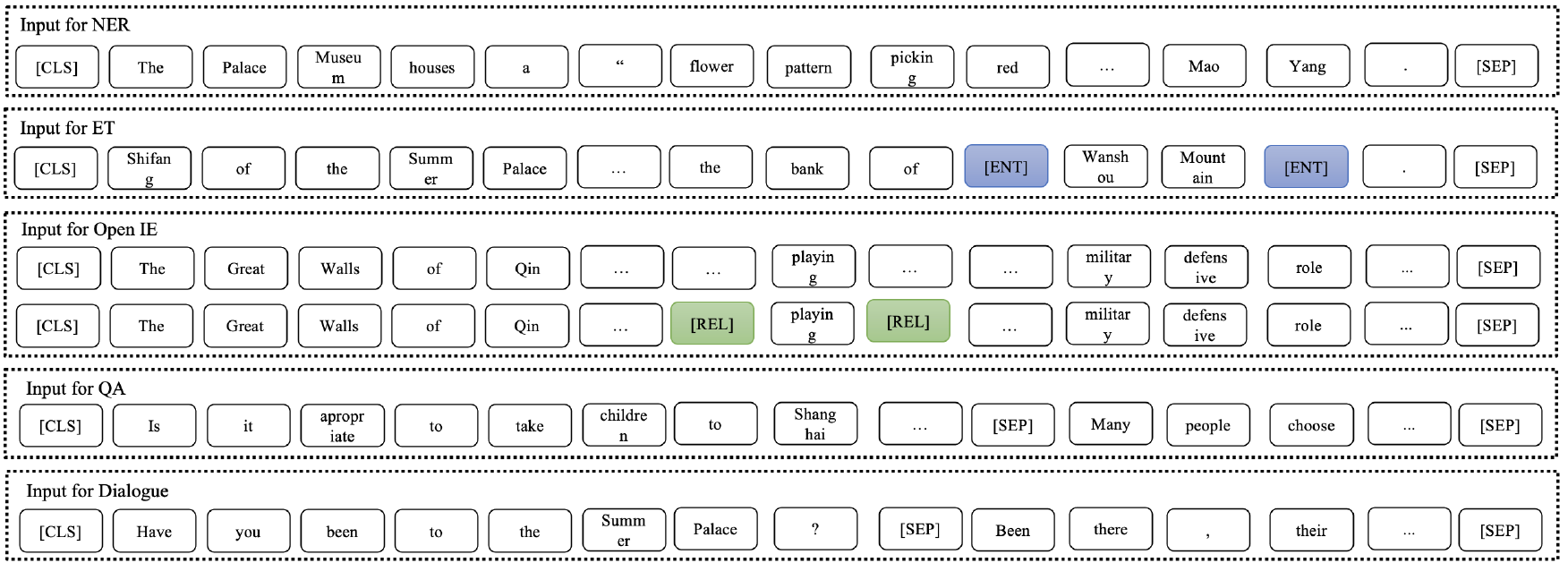}
\caption{The input scheme of tourism NLP tasks}
\label{input.fig}
\end{figure*}

The input form of each downstream tourism NLP task is shown in Figure \ref{input.fig}. When fine-tuning downstream tasks, our model does not need to change the input text because the model can learn heterogeneous knowledge in the pre-training stage and learn better parameters, like GPT-3 (\cite{DBLP:conf/nips/BrownMRSKDNSSAA20}) and WKLM (\cite{DBLP:conf/iclr/XiongDWS20}). Then the model uses the learned knowledge (parameters) to better solve downstream tasks.

For the NER task, we adopt the sequence labeling (\cite{DBLP:conf/naacl/LampleBSKD16}) scheme and use the vector of each token in the last layer to classify the entity labels.

The fine-grained entity typing (\cite{jin2019fine}) task aims to assign fine-grained type labels to the entity mention in the text. We add two special symbols [ENT] to highlight the entity mention and use the [CLS] vector of the last layer to classify the labels.

For the open IE task, we use a two-stage span extraction reading comprehension model (\cite{DBLP:conf/acl/LiYSLYCZL19,DBLP:conf/pakdd/LyuSLHLS21}). Specifically, we first train a relation prediction model to extract multiple predicate spans in the sentence. The way the model extracts each span is to predict the start and end positions. We use a threshold to select multiple spans, since each sentence may have more than one triple. We add two special symbols [REL] to highlight the predicate span. Then we train an entity prediction model to extract subject and object spans for each predicate.

For the QA task, we use the [CLS] vector in the last layer to calculate and rank the matching score of each candidate answer to the question.

For the dialogue task, we adopt a retrieval-based model. The training task is to predict whether a candidate is the correct next utterance given the context. For the test, we selected the candidate response with the largest probability. 
\section{Experiments}
\label{section4}
We perform experiments on the following tourism NLP tasks, including NER, Open IE, dialogue, QA, and fine-grained entity typing to evaluate our proposed models. Then we conduct ablation studies and analyze the influence of KG quality on the model. We then perform a significance test on the model results to mitigate the effects of random factors. Finally, we expand this method to the general domain and conduct experiments on the XNLI dataset. 

\subsection{Data and Setup}
We first introduce the data sources and statistics of the pretraining dataset. Then we introduce the statistics and annotation process of the downstream task datasets.
\subsubsection{Pre-training corpus}
Our pre-training corpus is composed of the plain text corpus and Baidu encyclopedia webpages of Chinese tourist attractions. We obtained the Chinese tourist attractions and fine-grained tourism types from ctrip.com, visitbeijing.com.cn, tripadvisor.cn, meituan.com, etc., and constructed a Chinese Tourism Knowledge Graph (CTKG). 
Then, we obtained 49,273 Baidu encyclopedia webpages based on the tourist attractions in the CTKG, including 0.27M knowledge triples describing tourist attractions. The plain text corpus (279M tokens) contains 174,326 travel guides collected from ctrip.com, mafengwo.cn, etc., and the plain text of 49,273 Baidu encyclopedia webpages of Chinese tourist attractions. 
We segment the document and limit the maximum length of the input text fragment to about 400 tokens so that the model can learn complex context. We convert the infobox into knowledge triples and then retrieve relevant triples for each text fragment in the same document. 

\subsubsection{Downstream Datasets}
For downstream tourism NLP tasks, due to the lack of sufficient evaluation datasets in the tourism domain, we adopt a well-known dialogue dataset KdConv\footnote{https://github.com/thu-coai/KdConv} (\cite{DBLP:conf/acl/ZhouZHHZ20}) and construct 4 tourism NLP datasets. We first use programs to annotate the raw data, then manually check and refine the labels. Table \ref{DataStatistics} lists the detailed information of the tourism NLP datasets for each task. We will introduce the process of labeling datasets for downstream tasks. 

For the tourism NER dataset, the raw data comes from the Baidu encyclopedia, and these data are also used to build a multimodal Chinese tourism knowledge graph\footnote{https://travel.xlore.cn/} (MCTKG) (\cite{DBLP:conf/icycsee/XieDWZJ0L21}). We use the 6 types of entities involved in MCTKG as dictionaries to obtain preliminary annotation results. Encoding issues with special characters can cause index shifts. A lack of understanding of entity context can lead to wrong entity types. Then, we manually add, delete, and modify unqualified labels. 

For the TravelQA dataset, we use a crawler to get user questions from the travel guide channel of ctrip.com. We select the answer adopted by the questioner as the gold answer from the replies, and the negative answers come from replies to other questions. 

For the TravelET dataset, we use the ontology of MCTKG for automatic labeling through the entity linking function of the XLORE platform (\cite{DBLP:journals/dint/JinLZHLZ19}). We then manually check and revise unqualified labels. 

For the TravelOIE dataset, we extract open relations from travel domain text by combining information extraction results based on dependency parsing (\cite{DBLP:conf/emnlp/QiuZ14}) with those based on a pre-trained model. The model is trained on the SpanSAOKE (\cite{DBLP:conf/pakdd/LyuSLHLS21}) dataset developed by Lyu et al., which is derived from the SAOKE (\cite{DBLP:conf/wsdm/SunLWFFL18}) dataset. Then, we manually add, delete, and modify unqualified triples, e.g. meaningless triples and semantically incomplete triples. 

\begin{CJK*}{UTF8}{gbsn}
\begin{table*}[!ht]
\caption{Dataset details}
\centering
\tiny
\resizebox{\linewidth}{!}{
    \begin{tabularx}{\linewidth}{c|p{1.0cm}XX}
    \toprule
    Dataset   & Train/Dev/Test & \multicolumn{1}{c}{Description} & \multicolumn{1}{c}{Example}  \\ 
    \midrule
    KdConv &  1,200\newline150\newline150 & This dataset is developed by (\cite{DBLP:conf/acl/ZhouZHHZ20}) through crowdsourcing to generate multi-turn conversations related to the domain-specific knowledge graph. Provided for public access from April 2020, it contains 3 domains, each with 1,500 dialogues, and the average number of turns is about 19. Considering the scope of this paper, we only use the travel domain dataset for evaluation. & A: 你去过颐和园吗？ (Have you been to the Summer Palace?)
    \newline B: 去过啊，他们的票价...这里几点可以玩啊？ (Been there, the fare is ... What time can I play here?)
    \newline A: 4月1日-10月31日是6:30-18:00开放;... (Open from 6:30-18:00 from April 1st to October 31st;...)
    \newline B: ... \\ \hline
    TravelQA & 5,400\newline675\newline675 & This is a travel domain question answering dataset. It is created based on ctrip.com. The questions are about traveling in different cities in China. The dataset comes from ctrip.com users’ questions and other users’ answers. Each question averagely contains 30 candidate answers, only one is the best answer, and the negative candidate answers are retrieved from the answers to all other questions based on BM25. It contains 6,750 questions and 209,234 QA pairs. & Q: 北京跟团游去香山公园看红叶的有没有推荐的团 和我妈妈一起的 (My mother and I are on a group tour in Beijing and want to visit Xiangshan Park to see the red leaves. Are there any recommended groups?) \newline
    (1) Now the leaves are not red, it will be better at the end of the month. There were too many people in Xiangshan when it was red...
    \newline(2) It is not recommended that you go to Sanya in August ... \\ \hline
    TravelNER  & 2,955\newline375\newline347 & This is a manually annotated tourism NER dataset. The sentences are taken from the Baidu encyclopedia webpages of Beijing tourist attractions. It includes 6 types of entities (Building, Attraction, Store, Person, Relic, and Organization). The entity vocabulary comes from the Chinese tourism KG we built. & \textit{故宫博物院藏，杨茂制造“花卉纹剔红渣斗”。} (\textit{The Palace Museum houses a "flower pattern picking red slag bucket" made by Mao Yang.}) \newline \textbf{Entity}: \textit{故宫博物院 (The Palace Museum):Attraction, 杨茂 (Mao Yang):Person, 花卉纹剔红渣斗 (flower pattern picking red slag bucket):Relic}.  \\ \hline
    TravelET & 168,679\newline18,697\newline4,586 & This is a fine-grained entity typing dataset in the tourism domain. The sentences are taken from the Baidu encyclopedia webpages of Beijing tourist attractions. The dataset contains 34 fine-grained tourist attraction labels which come from the Chinese tourism KG we built. The training set is automatically annotated based on anchor text and dictionary matching. The development and test sets are manually annotated. & \textit{北京颐和园石舫...位于昆明湖的西北部，\underline{万寿山}的西麓岸边。} (\textit{The Shifang of the Summer Palace in Beijing ... is located in the northwestern part of Kunming Lake, on the bank of the western foot of \underline{Wanshou Mountain}.}) \newline \textbf{Entity}: \textit{万寿山} (\textit{Wanshou Mountain})  --- /Attraction, /Attraction/NaturalScenery, /Attraction/NaturalScenery/WildScenery. \\ \hline
    TravelOIE & 21,416\newline2,677\newline2,223 & This is a tourism domain open IE dataset sampled from Baidu encyclopedia webpages of Beijing tourist attractions. We adopt ZORE (\cite{DBLP:conf/emnlp/QiuZ14}), a well-known Chinese open IE toolkit, to extract triples from the text as the gold standard in the training set. The development and test sets are manually annotated. & \textit{...战国秦、赵、燕长城仍在发挥军事防御作用，虽然...。} (\textit{... the Great Walls of Qin, Zhao, and Yan in the Warring States were still playing a military defensive role, although ...})
    \newline \textbf{Triple}: \textit{(战国秦、赵、燕长城,发挥,军事防御作用) ((the Great Walls of Qin, Zhao, and Yan in the Warring States, playing, a military defensive role))}. \\
    \bottomrule
    \end{tabularx}
}

\label{DataStatistics}
\end{table*}
\end{CJK*}

\subsection{Evaluation}
\begin{itemize}
	\item For the TravelQA dataset, the two metrics used to evaluate the quality of our model are Mean Average Precision (MAP) and Mean Reciprocal Rank (MRR) which are commonly used in the evaluation. MRR@N means that only the top-N candidates are considered. 
	\item For the TravelNER dataset, the evaluation metrics are micro-averaged precision, recall, and F1. 
	\item For the TravelET dataset, the evaluation metrics are accuracy, micro-F1, and macro-F1 score. 
	\item For the TravelOIE dataset, the evaluation metrics are micro-averaged precision, recall, and F1. 
	\item For the KdConv dataset, the evaluation metrics are hits@N, and Distinct-N. Hits@N denotes the rate that the correct answers are ranked in the top N. Distinct-1/2/3/4 (\cite{DBLP:conf/naacl/LiGBGD16}) is used to evaluate the diversity of generated responses.
\end{itemize}

\subsection{Hyper-parameters}
We use the pre-trained bert-base-Chinese\footnote{https://huggingface.co/bert-base-chinese} as the baseline to further train the language model. The learning rate is 3e-5. Each input text sequence contains about 400 tokens. We use the AdamW (\cite{loshchilov2017decoupled}) optimization algorithm to update the model parameters. For downstream tasks, we run the program 3 times and take the median as the final result. 

We list the main hyper-parameters for downstream tasks. For TravelNERT, we set learning rate = 8e-5, batch size = 64, epoch = 40. For TravelFET, we set max sequence length = 256, batch size = 256, learning rate = 2e-5, epoch = 3, threshold = 0.3, warmup proportion = 0.2. For TravelOIE, predicate extraction stage sets max sequence length = 150, training epoch = 2, learning rate = 2e-5. Entity extraction stage sets max sequence length = 150, training epoch = 3, learning rate = 3e-5. For TravelQA, we set batch size = 256, learning rate = 2e-5, epoch = 10, save step = 500. For KDConv, we set epoch = 3, warmup proportion = 0.1, batch size = 8, learning rate = 5e-5. 

These experiments are run on an Intel(R) Xeon(R) Platinum 8163 CPU @ 2.50GHz (Mem: 330G) with 8 Tesla V100s (32G) and an Intel(R) Xeon(R) CPU E5-2680 v4 @ 2.40GHz (Mem: 256G) with 8 RTX 2080Tis (11G). We use batch size=2048, and conduct a 1-day pre-training on the free text corpus and multi-format text corpus respectively.

\subsection{Results on The Tourism NLP Tasks}

\begin{table*}[]
\caption{Results on the 5 downstream tourism NLP datasets}
\resizebox{\linewidth}{!}{
\begin{tabular}{c|ccc|ccc|ccc|ccc|cccccc} \toprule
                 & \multicolumn{3}{c|}{TravelNER} & \multicolumn{3}{c|}{TravelET} & \multicolumn{3}{c|}{TravelOIE} & \multicolumn{3}{c|}{TravelQA} & \multicolumn{6}{c}{KDConv}                                          \\
Metrics          & P        & R       & F1       & Acc    & Mi-F1    & Ma-F1    & P        & R       & F1       & MAP    & MRR@5    & MRR@1    & Hits-1 & Hits-3 & Dist-1 & Dist-2 & Dist-3 & Dist-4 \\ \midrule
Baidu-ERNIE      &29.9 &32.9 & 31.3        &  {\bf 63.7} & 72.4 & 61.7   &  39.2 & {\bf 30.6} & 34.4   &     85.0   &     84.4     &    77.8      &   {\bf 49.7} & {\bf 76.3} & {\bf 7.4} & {\bf 23.5} & {\bf 35.7} & {\bf 43.0}      \\
K-BERT(TravelKG) & 50.5 &58.3 &54.1        &     --   &    --      &   --       &      --    &    --     &   --       &   82.5     &     81.6     &  75.0        &  --      &   --     &      --      &      --      &     --       &      --      \\
BERT-wwm & 49.1 & {\bf 64.4} & 55.7       &    62.3     &      71.9    &    61.9       &   34.6      &   20.1     &     25.4     &    80.3    &     79.4     & 72.0        &   45.1     &    73.9     &       7.1     & 22.7    &     34.4        &    41.3      \\
BERT\textsubscript{BASE}             & 45.2 & 61.5 & 52.1         & 63.5 & 72.5 & 62.6      &     38.8 & 29.7& 33.6   &    82.4    &     81.5     &    74.5      &     45.3 & 71.9&  { 7.2}&22.6 & 34.0& 40.8        \\ \midrule
\textbf{TravelBERT\textsubscript{C}}     &48.2 & 60.4 & 53.6          &  63.6 & 72.3 &   62.2       &  39.5 & 30.5& 34.4    &     84.4   &    83.7      &    77.5      & 41.5 & 69.3&  6.9& 21.4& 32.0&  38.2       \\
\textbf{TravelBERT\textsubscript{K}}     & \textbf{50.9} & { 62.3} &  {\bf 56.0}  &  63.6 & {\bf 73.3}&  {\bf 63.4}     &    {\bf 39.9}&  30.5 & {\bf 34.6}     &   \textbf{85.2}     &    \textbf{84.7}      &   \textbf{78.4}       &   45.5 & 72.7& 7.2  & 22.7 & 34.3 &  41.3   \\ \bottomrule       
\end{tabular}
}

\label{5nlpres}
\end{table*}

Table \ref{5nlpres} lists the results on the 5 tourism NLP datasets. BERT\textsubscript{BASE} represents the common pre-trained Chinese BERT model. TravelBERT\textsubscript{C} and TravelBERT\textsubscript{K} represent the use of plain text and HKLM to further pre-train the language model, respectively. K-BERT(TravelKG) denotes that the K-BERT (\cite{DBLP:conf/aaai/LiuZ0WJD020}) model uses our Chinese tourism KG for training and prediction. Baidu-ERNIE (\cite{DBLP:journals/corr/abs-1904-09223}) represents that we use the Chinese version ERNIE 1.0\footnote{https://huggingface.co/nghuyong/ernie-1.0}. BERT-wwm\footnote{https://huggingface.co/hfl/chinese-bert-wwm} (\cite{chinese-bert-wwm}) represents the model trained using the whole word masking (wwm) strategy.

For the TravelNER dataset, TravelBERT\textsubscript{C} can increase the precision score by +1.5\%, while the recall slightly degrades. This shows that the use of unstructured text in a specific domain to further pre-train the language model is conducive to a more accurate understanding of entity concepts because the contextual description of related entities is richer. TravelBERT\textsubscript{K} achieves the best results, indicating that pre-training with entity-centric heterogeneous resources is helpful for this task. 
Both structured and unstructured data in a specific domain can help identify domain-specific entities. Although the amount of encyclopedia pre-training corpus is 1/4 of the plain text pre-training corpus, TravelBERT\textsubscript{K} achieves better results than TravelBERT\textsubscript{C}. This demonstrates that heterogeneous knowledge is effective in the pre-training process.



For the TravelET dataset, we can see that the performance of TravelBERT\textsubscript{C} is almost identical to the baseline, which means simply using unstructured text to further pre-train the language model may not bring significant improvements to this task. TravelBERT\textsubscript{K} achieves improvements by +0.8\% in micro-F1 and macro-F1, which indicates that the entity typing task requires entity-centric heterogeneous knowledge. 



For the TravelOIE dataset, our TravelBERT\textsubscript{C} and Travel-BERT\textsubscript{K} outperform the baseline by +0.8\% and +1\% micro-F1 respectively. This demonstrates that heterogeneous knowledge is beneficial to open information extraction tasks.


For the TravelQA dataset, we observe that TravelBERT\textsubscript{C} improves the MAP score by +2\% and improves MRR@N by more than +2\%. This means that pre-training with domain-specific unstructured text is helpful for this task. TravelBERT\textsubscript{K} improves the MAP score by +2.8\%. Baidu-ERNIE also achieves an identical result, which means that the model is good at handling Chinese question answering tasks. 


KDConv dataset is a knowledge-driven conversation dataset in which each response is generated based on a specific triple. Fine-tuning the BERT with unstructured text degrades the results. This means for knowledge-driven conversation tasks, simple pre-training with unstructured text in a specific domain may hurt performance. This is because, for knowledge-driven dialogue, the model may not be able to efficiently utilize the unstructured context. Using the proposed HKLM can improve hits-N and Distinct-N scores. This means that TravelBERT\textsubscript{K} can inject the travel knowledge of interest into the conversation. Baidu-ERNIE achieves the best results, which means that the model is good at handling Chinese dialogue tasks. This is because the data used by Baidu-ERNIE in the pre-training process is many times that of ours, including the multi-round dialogue data of Baidu Tieba. This also means that pre-training with more data, especially data related to the target task, can improve the performance of the model on the target task.

We observe that the proposed HKLM achieves significant improvements on the TravelNER and TravelQA datasets, and has minor improvements on other datasets. This is because knowledge triples can enhance the representation of entities in the TravelNER dataset. Learning paragraph semantics helps understand the TravelQA dataset. Compared with TravelBERT\textsubscript{C}, the HKLM also greatly improves the results of the KDConv dataset which requires knowledge triples to return the correct information response. Nevertheless, since the labels of the TravelET dataset have a logical hierarchy, there is a lack of understanding of the hierarchical structure of classes (taxonomy) in the pre-training process. In addition, the task is a multi-label classification problem, and it is relatively simple to use the threshold to select the final labels. For the TravelOIE dataset, the data annotation relies on the information extraction mechanism 
of dependency parsing (\cite{DBLP:conf/emnlp/QiuZ14}), but we did not specifically add linguistic knowledge during the pre-training process. These issues need to be further explored in the future.
\subsection{Ablation Study}
We perform ablation studies on the TravelNER and TravelQA datasets respectively because these two datasets can reflect the entity-oriented task and paragraph-oriented task. Then, we analyze the influence of knowledge triples, titles, and KG quality in the pre-training process.

\begin{table}[!htbp]
\caption{Ablation results on the TravelNER dataset}
\small
\centering
    \begin{tabular}{l|ccc}
    \toprule
    Settings          & P  & R & F1    \\ 
    \midrule
    TravelBERT\textsubscript{K} & {\bf 50.9} &  62.3 &  {\bf 56.0} \\
    --titles &49.3 &63.9 &55.6 \\
    --triples &50.8 &60.4 &55.2 \\
    --triples, titles &46.5&{\bf 64.2}&53.9 \\
    \bottomrule
    \end{tabular}

\label{ablNER}
\end{table}

As shown in Table \ref{ablNER}, we observe that removing the titles slightly reduces the F1 by -0.4\%. When we remove the knowledge triples, the F1 drops by -0.8\% because knowledge triples help learn the entity knowledge. This means that heterogeneous knowledge is beneficial for the TravelNER task. Comparing Table \ref{ablNER} and Table \ref{5nlpres}, we can see that removing both triples and titles gives worse results than TravelBERT\textsubscript{C}. This is because TravelBERT\textsubscript{K} only uses the Baidu encyclopedia corpus to pre-train the model while TravelBERT\textsubscript{C} uses the Baidu encyclopedia corpus and travel guides. This means that the size of the pre-training corpus and the content of the corpus have an impact on the pre-training results. 

\begin{table}[!htbp]
\caption{Ablation results on the TravelQA dataset}
\centering
\small
    \begin{tabular}{l|cccc}
    \toprule
    Settings   & MAP    & MRR@10 & MRR@5 & MRR@1   \\ 
    \midrule
TravelBERT\textsubscript{K} & {\bf 85.2} &{\bf 85.0} &{\bf 84.7} & {\bf 78.4} \\
 --titles &84.0 &83.8 &83.3 &76.6 \\
 --triples &83.8 &83.5 &83.3 &76.3 \\
    --triples, titles & 83.0 & 82.8 & 82.4 & 75.1 \\ 
    \bottomrule
    \end{tabular}

\label{ablQA}
\end{table}

Removing the titles degrades the MAP score by -1.2\%, as shown in Table \ref{ablQA}. This means that titles benefit the TravelQA dataset. After removing the knowledge triples, the performance drops by -1.4\%. This means that titles and knowledge triples help understand the paragraph. When we remove both triples and titles, the model becomes TravelBERT\textsubscript{C} which uses part of the free text for pre-training. 

\begin{table}[!htbp]
\caption{Results on the TravelNER dataset after pre-training with KGs of different quality}
\small
\centering
    \begin{tabular}{l|ccc}
    \toprule
    Settings          & P  & R & F1    \\ 
    \midrule
    TravelBERT\textsubscript{K} & {\bf 50.9} &  {\bf 62.3} &  {\bf 56.0} \\
    --50\% triples &47.9 &62.3 &54.2 \\
    +noise &45.8 &62.0 &52.7 \\
    \bottomrule
    \end{tabular}

\label{noisener}
\end{table}
In addition to directly removing KG, we further explore the impact of the KG quality on the pre-training process. We adopt two methods to reduce the KG quality. First, we randomly remove some knowledge triples (--50\%) to simulate an incomplete knowledge graph. We observe that the result drops by -1.8\% in the F1 score, as shown in Table \ref{noisener}. Second, we modify the knowledge triples (+noise) by randomly masking the attribute values to simulate a noisy KG. We observe that the F1 score drops by -3.3\%. This means that the deterioration of the KG quality will hurt the pre-training process. 

\subsection{Significance Test}
We perform a significance test on the model results to mitigate the effects of random factors. The t-test (\cite{allua2009inferential}) is an inferential statistic used to determine whether there is a significant difference between the mean values of two groups. We run the experiment 10 times to generate the F1 score. As shown in Table \ref{signtest}, columns 1,2,...,10 represent the experimental rounds. As can be seen from the mean values in Table \ref{signtest}, our HKLM improves the F1 score. Table \ref{mixsig} lists the p-values for t-tests for pair-wise models. Let the significance level $\alpha$ = 0.05, and the null hypothesis will be rejected when the p-value is less than $\alpha$ (\cite{xirunchen2009}). As shown in Table \ref{mixsig}, we bold p-values less than $\alpha$. We found that when the difference between the means of the two groups of data is greater than 0.8, the p-value will be less than $\alpha$, that is, the null hypothesis with a confidence level of 1-$\alpha$ is rejected. We can claim at 95\% confidence level that the proposed method achieves better results than other models in this experiment. With pairwise comparisons of all models, we can also see that the results of many methods may not be statistically significantly different. However, differences in mean, max and min values of different models can still reflect their differences. This also demonstrates the importance of model selection for real-world system performance. 
\begin{table*}[!htbp]
\caption{F1 scores on the TravelNER dataset}
\small
\centering
    \begin{tabular}{l|ccccccccccc}
    \toprule
    Models          & 1  & 2 & 3 & 4  & 5 & 6 & 7  & 8 & 9 & 10 & Mean \\ 
    \midrule
     Baidu-ERNIE & 32.3 & 32.0  & 32.3 & 35.0 & 30.6  & 27.3 & 32.1 & 33.8  & 32.5 & 27.7 & 31.6 $\pm$ 2.3 \\
     K-BERT(TravelKG) & 54.8 &  54.5 & 52.6 & 55.3 & 56.2  & 53.7 & 56.5 & 53.4  & 54.1 & 53.9 & 54.4 $\pm$ 1.2 \\
     BERT-wwm & 56.0 &  54.9 & 55.5 & 56.2 &  53.9 & 56.1 & 54.9 & 55.0  & 55.3 & 53.9 & 55.2 $\pm$ 0.8 \\
     BERT\textsubscript{BASE} & 55.6 & 55.2  & 54.9 & 56.6 & 55.5  & 52.8 & 55.4 &  53.9 & 54.5 & 54.1 & 54.9 $\pm$ 1.0 \\ \midrule
    TravelBERT\textsubscript{C} & 56.0 & 54.3  & 54.3 & 55.9 & 56.1  & 56.2 &53.9 & 54.5  & 56.4 & 56.2 & 55.3 $\pm$ 0.9 \\
    TravelBERT\textsubscript{K} & 55.8 & 56.0  & 57.5 & 56.5 & 55.8  & 56.0 & 56.6 & 55.2  & 55.5 & 54.9 & {\bf 56.0} $\pm$ {\bf 0.7}\\
    \bottomrule
    \end{tabular}

\label{signtest}
\end{table*}

\begin{table*}[!htbp]
\caption{P-values for t-tests for model pairs}
\small
\centering
    \begin{tabular}{l|cccccc}
    \toprule
    Models          & Baidu-ERNIE  & K-BERT(TravelKG) & BERT-wwm & BERT\textsubscript{BASE}  & TravelBERT\textsubscript{C} & TravelBERT\textsubscript{K}  \\ 
    \midrule
     Baidu-ERNIE & 1.0  & {\bf 6.6e-16} & {\bf 1.4e-16} & {\bf 3.2e-16}  & {\bf 1.8e-16}  & {\bf 6.6e-17}  \\
     K-BERT(TravelKG) & {\bf 6.6e-16}  & 1.0 & 0.170 & 0.505  & 0.095 & {\bf 0.004}  \\
     BERT-wwm & {\bf 1.4e-16}  & 0.170 & 1.0 & 0.463  & 0.613 & {\bf 0.033}  \\
     BERT\textsubscript{BASE} & {\bf 3.2e-16}  & 0.505 & 0.463 & 1.0  & 0.265 & {\bf 0.013}  \\ 
    TravelBERT\textsubscript{C} &  {\bf 1.8e-16} & 0.095 & 0.613 & 0.265  & 1.0 & 0.144  \\
    TravelBERT\textsubscript{K} & {\bf 6.6e-17}  & {\bf 0.004} & {\bf 0.033} & {\bf 0.013}  & 0.144 & 1.0 \\
    \bottomrule
    \end{tabular}

\label{mixsig}
\end{table*}

\subsection{Results of HKBERT on XNLI Dataset}
We further expand to domain-agnostic HKBERT by pre-training on all Baidu encyclopedia articles. The training data consists of over 20 million Baidu encyclopedia web pages (49GB of plain text data), and triples extracted from unstructured paragraphs using open relation extraction. Open relation extraction enriches the content of structured data, and finally we jointly pre-train unstructured, semi-structured, and well-structured text.

We evaluate the pre-trained HKBERT on the textual entailment task. Textual entailment is an important benchmark task in natural language understanding, which aims to classify the relationship between two sentences as entailment or contradiction or neutral. The XNLI dataset (\cite{DBLP:conf/emnlp/ConneauRLWBSS18}) is a multi-domain cross-language large-scale textual entailment dataset proposed by Facebook in 2018. The research scope of this paper is a Chinese pre-trained language model, so only the Chinese dataset is used for experiments. The number of samples for training/development/test sets is 392,702/2,490/5,010 respectively.

Table \ref{xnli} lists the experimental results of multiple Chinese pre-trained language models fine-tuned on the XNLI Chinese dataset. BERT (\cite{DBLP:conf/naacl/DevlinCLT19}) is the initialized baseline model used in this paper. ERNIE 1.0 (\cite{DBLP:journals/corr/abs-1904-09223}) is proposed by Sun et al. to enhance masked language models by exploiting word, entity, and phrase level masks. BERT-wwm (\cite{chinese-bert-wwm}) is proposed by Cui et al. and contains 3 different masking strategies, including whole word masking, N-gram masking, MLM as correction masking. BERT-wwm-ext, RoBERTa-wwm-ext, and RoBERTa-wwm-ext-large represent these models using a larger pre-trained corpus or more model parameters. It can be seen that HKBERT achieves an accuracy of 79.0\%, which is 1.2\% higher than the original BERT, and also higher than the RoBERTa-wwm-ext. RoBERTa-wwm-ext-large achieves the highest performance, but the model has several times the parameters of HKBERT and also uses a larger pre-trained corpus. Experimental results demonstrate that the proposed method helps to improve the performance in the general domain. 

\begin{table}[!htbp]
\centering
\small
\caption{Experimental results on the XNLI dataset}
    \begin{tabular}{l|c}
    \toprule
    Model   & Accuracy(\%)      \\ 
    \midrule
BERT  & 77.8  \\
ERNIE 1.0 & 78.6 \\
BERT-wwm & 78.2  \\
BERT-wwm-ext & 78.7 \\
RoBERTa-wwm-ext & 78.8 \\
RoBERTa-wwm-ext-large & \textbf{81.2}  \\ \midrule
HKBERT & 79.0  \\
    \bottomrule
    \end{tabular}
\label{xnli}
\end{table}

\section{Conclusion and Future Work}
\label{section5}
This paper presents a pre-training approach incorporating unstructured, semi-structured, and well-structured text in the same contextual representation space. Specifically, the proposed HKLM models the document structure, relevant knowledge triples, and plain text and realizes the interaction between heterogeneous knowledge. We construct 4 tourism NLP datasets, and the experimental results show that the further use of multi-format text resources in pre-training can help improve the performance of downstream tasks. We expand this approach to the general domain and demonstrate that performance improvements can be achieved on the XNLI dataset. This paper mainly resolves the problem of finding the correspondence and unified representation between multi-format text resources in the pre-training process. The innovation of this approach lies in that we formulate this problem by using entity-oriented heterogeneous knowledge resources and corresponding learning algorithms. In the future, we plan to further improve the semantic awareness of this method and generalize it comprehensively to the general domain.

\section*{Acknowledgements}
This work is supported by the Key-Area Research and Development Program of Guangdong Province (2019B010153002), a grant from the Institute for Guo Qiang, Tsinghua University (2019GQB0003), and Huawei Noah's Ark Lab.










\bibliographystyle{cas-model2-names}

\bibliography{cas-refs}



\bio{}
\endbio

\bio{}
\endbio

\end{document}